\documentclass[letterpaper, 10 pt, journal, twoside]{IEEEtran}

\usepackage{amsmath,amsfonts,amssymb}
\usepackage{mathtools}
\usepackage{array}

\usepackage[labelformat=simple]{subfig}
\usepackage[font={footnotesize}]{caption}
\usepackage{marginnote}
\usepackage{textcomp}
\usepackage{verbatim}
\usepackage{graphicx}
\usepackage{cite}
\usepackage{gensymb}
\usepackage{booktabs}
\usepackage{float}
\usepackage{mathrsfs}
\usepackage{bm}
\usepackage{makecell}
\usepackage{color,xcolor}
\usepackage{babel}
\usepackage{booktabs}
\usepackage{threeparttable}
\usepackage{url}
\usepackage{diagbox}
\usepackage{stfloats}
\usepackage{amsopn}
\usepackage{multirow}
\usepackage{ulem}
\usepackage{enumitem}
\usepackage{setspace}

\usepackage[
            citecolor=black,
            ]{hyperref}

\allowdisplaybreaks[4]

\title{
Hybrid Trajectory Optimization for Autonomous Terrain Traversal of Articulated Tracked Robots}
\vspace{-0.7cm}
\author{Zhengzhe Xu$^{1}$, Yanbo Chen$^{2}$, Zhuozhu Jian$^{2}$, Junbo Tan$^{2}$, Xueqian Wang$^{2}$, and Bin Liang$^{2}$%
\thanks{Manuscript received: May, 25, 2023; Revised October, 17, 2023; Accepted November, 15, 2023.}
\thanks{
This paper was recommended for publication by Editor Ashis Banerjee upon evaluation of the Associate Editor and Reviewers' comments.
This work was supported by the National Natural Science Foundation of China under Grant 62293545 and Grant 62103225. 
\textit{
(Zhengzhe Xu and Yanbo Chen are co-first authors.) 
(Corresponding authors: Junbo Tan; Xueqian Wang.)}
}
\thanks{$^{1}$Zhengzhe Xu is with the School of Mechanical Engineering and Automation, Harbin Institute of Technology, Shenzhen, China {\tt\footnotesize200320314@stu.hit.edu.cn}. 
}
\thanks{$^{2}$Yanbo Chen, Zhuozhu Jian, Junbo Tan, Xueqian Wang, Bin Liang are with the Center for Artificial Intelligence and Robotics, Shenzhen International Graduate School, Tsinghua University, Shenzhen, China 
{\tt\footnotesize\{cyb23@mails,
jzz21@mails,
tjblql@sz,
wang.xq@sz,
liangbin@mail\}.tsinghua.edu.cn}. }%
\thanks{Digital Object Identifier (DOI): see top of this page.}
}

\begin{document}

\maketitle
\begin{abstract}
Autonomous terrain traversal of articulated tracked robots can reduce operator cognitive load to 
enhance task efficiency and facilitate extensive deployment.
We present a novel hybrid trajectory optimization method aimed at generating efficient, stable, and smooth traversal motions. 
To achieve this, we develop a planar robot-terrain contact model and divide the robot's motion into hybrid modes of driving and traversing. 
By using a generalized coordinate description, the configuration space dimension is reduced, which facilitates real-time planning.
The hybrid trajectory optimization is transcribed into a nonlinear programming problem and divided into subproblems to be solved in a receding-horizon planning fashion. 
Mode switching is facilitated by associating optimized motion durations with a predefined traversal sequence. 
A multi-objective cost function is formulated to further improve the traversal performance.
Additionally, map sampling, terrain simplification, and tracking controller modules are integrated into the autonomous terrain traversal system.
Our approach is validated in simulation and real-world scenarios with the Searcher robotic platform. 
Comparative experiments with expert operator control and state-of-the-art methods show advantages in terms of time and energy efficiency, stability, and smoothness of motion.
\end{abstract}
\vspace{-0.1cm}
\vspace{-0.4cm}
\section{Introduction}
\label{sec:Introduction}
\begin{figure}[!t]
    \centering
    \includegraphics[width = 1.0\linewidth]{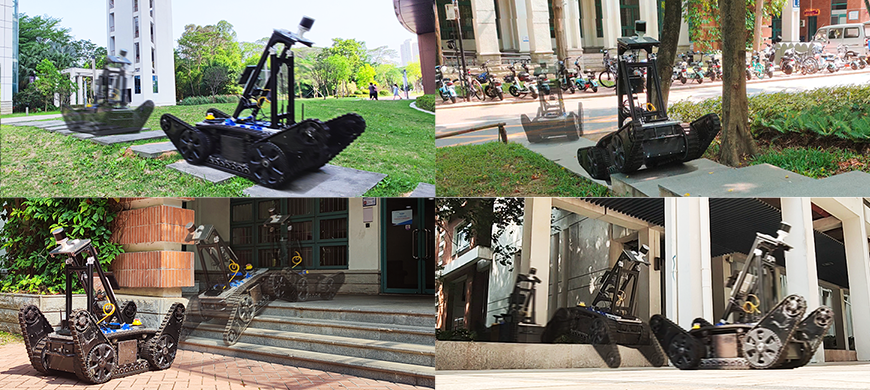}
    \caption{
    The \textit{Searcher}, an articulated tracked robot, can drive through terrain with minor undulations
    as well as utilize flippers to support itself, 
    enabling it to surmount uneven terrain such as high platforms and stairs with ease.
    }
    \label{fig:main_fig}
    \vspace{-0.6cm}
\end{figure}

\setstretch{0.95}
\IEEEPARstart{M}{OBILE} robots are revolutionizing a wide range of industries, from emergency response and disaster relief to environmental exploration and infrastructure maintenance \cite{klamt2019flexible,michael2012collaborative,niroui2019deep}. 
However, their successful deployment is heavily dependent on the ability to traverse challenging terrain and overcome obstacles with ease. 
Articulated tracked robots have emerged as a promising solution to these challenges, 
with hybrid locomotion: they can drive like a wheeled vehicle, with a large track contact area providing unparalleled stability; in addition, controllable flippers are used to actively maintain contact with the ground to traverse faulted terrain.
The Searcher that we use is an example as shown in Fig.\ref{fig:main_fig}. 

Articulated tracked robots are commonly operated through remote control, a demanding task that requires skilled operators to simultaneously monitor their surroundings while controlling the acceleration, steering, and flipper angles, making stability and efficiency an inherent trade-off.
A more efficient terrain traversal strategy is to maintain a certain speed while using an appropriate flipper configuration for support, which utilizes all degrees of freedom (DOFs) simultaneously rather than separated motions.
Autonomous execution of such optimized motions offers several benefits, including reducing the cognitive load on human operators and improving task efficiency. 
Optimization-based planning methods are particularly effective in generating efficient motion.
However, the hybrid locomotion introduced by flippers and the complex contact dynamics arising from the track's contact-rich nature pose significant planning challenges.
To overcome the challenges, we propose a planar simplified model to capture the contact between the robot and the terrain, optimizing the trajectory based on a predefined traversal sequence to achieve hybrid mode switching, resulting in efficient, stable, and smooth motion.

\vspace{-0.3cm}
\subsection{Related Work}
\label{sub:related}
Autonomous terrain traversal of articulated tracked robots is a long-studied problem. 
Early studies\cite{steplight2000mode,ohno2007semi,okada2011shared} have shown that preset simple strategies relying on sensor-reflection can achieve this goal.
However, these strategies are limited in generating complex motions and may not be applicable to different types of terrain.
In recent years, learning-based approaches have been used to address flipper control and autonomous traversal, such as neuroevolution \cite{sokolov2017hyperneat} and reinforcement learning (RL) \cite{pecka2016autonomous,mitriakov2021reinforcement}. 
However, these studies are conducted solely on a single type of terrain, while our work focuses on a general terrain traversal scheme. 

Many studies emphasize the significance of flipper configuration in terrain traversal and define discrete flipper templates (or shapes, states, postures) that effectively address configuration selection in different terrains. 
A D*-Lite-based path planning method is proposed in \cite{colas20133d}, where robot position planning and flipper configuration planning are separated. 
Suitable flipper configurations are selected among the four predefined flipper postures based on the terrain information on the path.
Similarly, a linear classifier is trained in \cite{zimmermann2014adaptive} to predict the appropriate shape for the terrain among five predefined flipper shapes.
In \cite{azayev2022autonomous}, seven states are defined, and a learnable soft-differentiable neural state machine architecture is proposed that considers the probability of switching between states.
However, these approaches do not consider overall motion coherence, which may lead to inefficient motion and high energy consumption due to frequent state switching. 
In contrast, a continuous traversal sequence is proposed in \cite{ohno2007semi}, which enabled cooperation between the front and rear flippers but lacked a unified strategy and clear criteria for specific phases. 
A simplified robot skeleton model is developed in \cite{yuan2019configuration}, and flipper motion is generated by searching a reduced-dimensional configuration space. 
However, it prioritizes the transitions between adjacent configurations over overall motion efficiency, and coordinate selection is not intuitive.

Optimization-based approaches excel in generating complex motions. 
A recent study \cite{chengeometry} proposes a
geometry‐based flipper motion planning (GFMP) method, 
using dynamic programming to optimize flipper motion. 
This method exhibits high terrain adaptability and has better stability and smoothness compared to RL-based methods.  
However, it can only perform low-resolution online planning in a limited horizon due to its high computational demands and lack of flexibility in imposing actuator constraints.
Trajectory optimization (TO) techniques can effectively handle constraints, and be effective in enhancing the agility of legged robots \cite{winkler2018gait}
as well as achieving kinematically feasible, statically stable motion for walking excavators in rough terrain \cite{jelavic2019whole}. 
Furthermore, the hybrid trajectory optimization (HTO) method has been utilized to generate multi-phase motions for hybrid systems with discontinuous dynamics \cite{posa2014direct, bjelonic2020rolling}.

For our Searcher robot, which weighs over 200kg and has a considerable contact area providing sufficient friction, we focus on kinematically feasible solutions to maintain static equilibrium.
However, the contact-rich nature presents a significant challenge in developing a robot-terrain contact (RTC) model suitable for planning purposes.
Previous studies analyze the interaction between the robot and terrain \cite{wang2008track, liu2009track}, but the model complexity impedes effective planning.

\subsection{Contributions}
    This work offers the following contributions:
    \begin{itemize}
        \item We propose a planar RTC model for articulated tracked robots to simplify the contact patterns. The configuration space is reduced to three dimensions by generalized coordinates, facilitating real-time planning capabilities.
        
        \item We develop a novel HTO formulation with a multi-objective cost function to generate hybrid motion and optimize trajectory in real-time with a receding horizon.
        
        \item We validate the integrated terrain traversal system in both simulation and real-world scenarios. 
        Comparisons with expert operator control and the state-of-the-art autonomous method demonstrate the advantages of our approach in terms of motion time and energy efficiency, stability, and smoothness. 
    \end{itemize}

\setstretch{1.0}
\vspace{-0.1cm}
\section{Modeling}
\label{sec:modeling}
\vspace{-0.1cm}
\setstretch{0.98}
\subsection{Planar Robot-Terrain Contact Model}
\vspace{-0.1cm}
\label{subsec:Robot and Terrain Modeling}
The Searcher has coaxial left and right flippers, indicating its suitability for traversing terrain with little undulation in the longitudinal section. 
Therefore, the RTC model is developed in the longitudinal plane to obtain a reduced configuration description through the contact between a simplified terrain and the robot model.
\subsubsection{Simplified Terrain Model}
\label{subsubsec:Simplified Terrain Model}
\begin{figure}[!t]
    \centering
    \subfloat[]{
        \includegraphics[width = 0.61\linewidth]{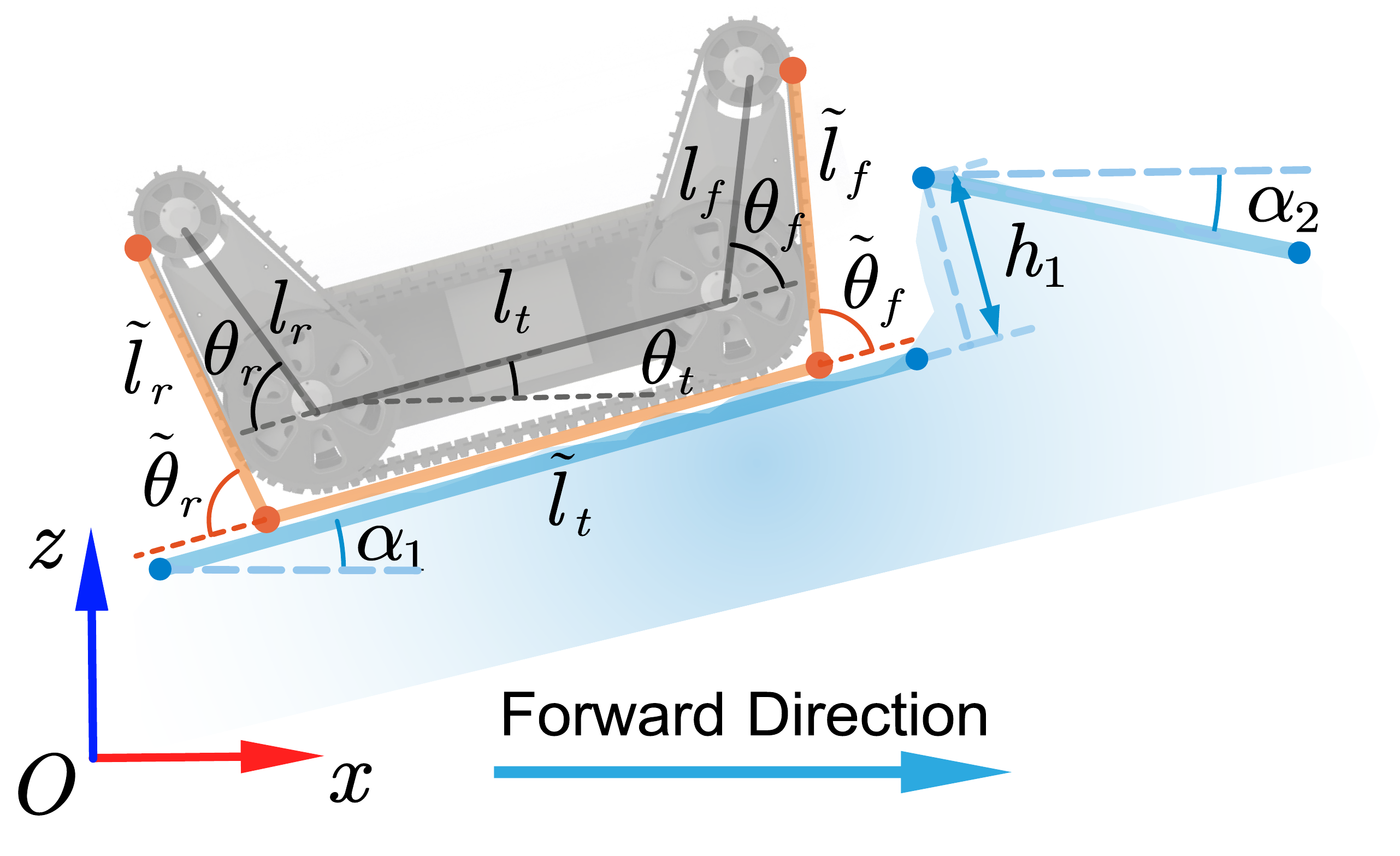}
        \label{fig:planar_model}
    }
    \subfloat[]{
        \includegraphics[width = 0.34\linewidth]{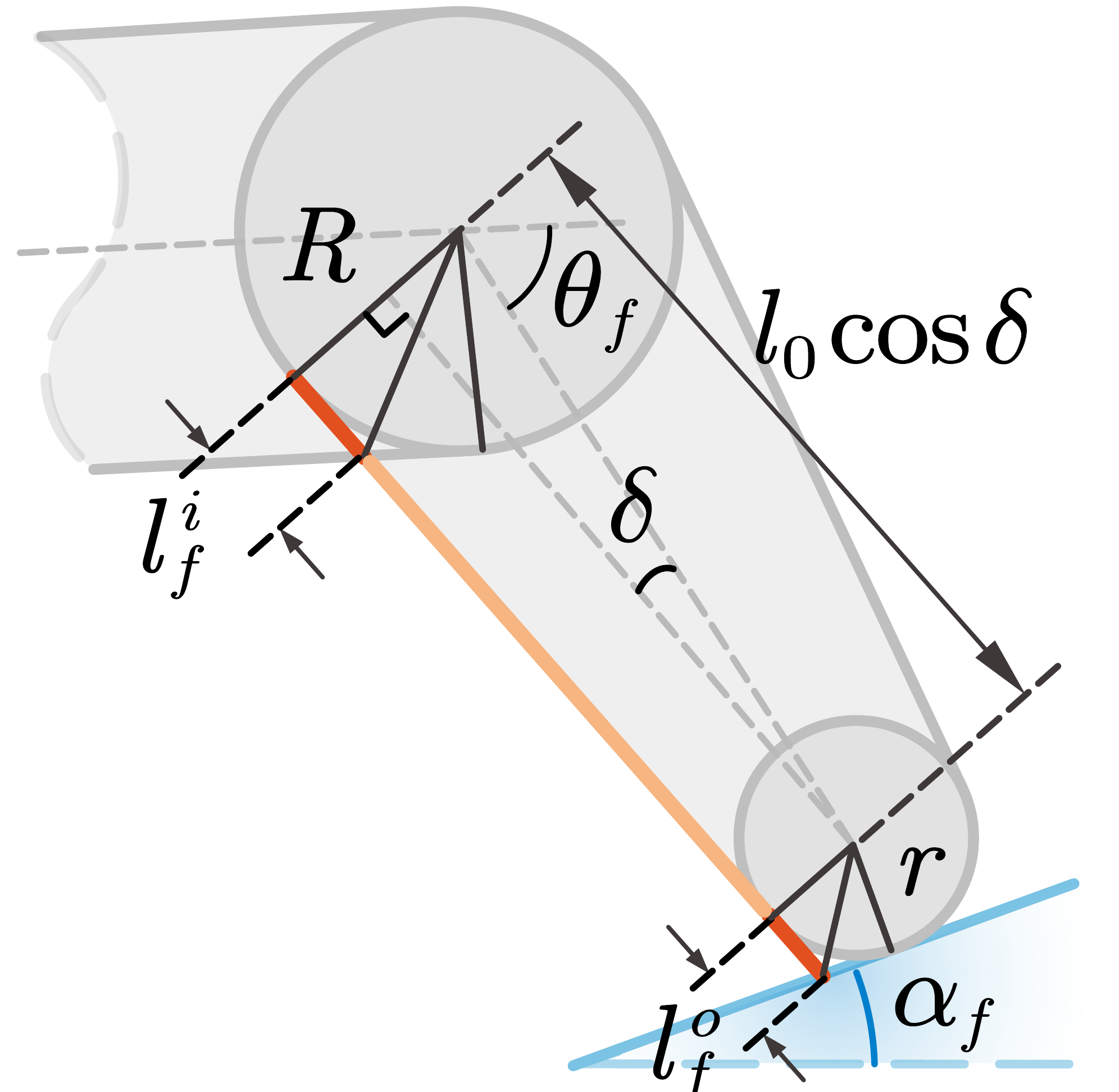}
        \label{fig:contact_approx}
        }
    \vspace{-0.15cm}
    \caption{Illustration of the simplified models. The flipper length $l_f=l_r=l_0$ and the track length $l_t$ are indicated in Fig. \ref{fig:planar_model}. $\theta_f$ and $\theta_r$ denote the front and rear flipper angles, $\theta_t$ is the pitch angle (set the robot head up as positive). The symbols with tilde indicate the model parameters. Fig. \ref{fig:contact_approx} shows the front flipper as an example of a model flipper length consisting of a fixed-length part $l_0\cos\delta$ and two variable-length parts $l_f^i$ and $l_f^o$. }
    \label{fig:modelling}
    \vspace{-0.6cm}
\end{figure}

Due to the large contact area provided by the track, the robot can traverse terrains with low undulation and unevenness without relying on the support of flippers. 
This insight motivates us to simplify the terrain into simple planes that are represented in two dimensions by line segments as shown in Fig. \ref{fig:planar_model}. The simplified terrain set is defined as $\mathbb{T} = \left\{ \tau _i \right\} _{i=1}^{N_t}$, where the terrain $\tau_i$ is described by the start and end point of the line segment. 
Each terrain can be parameterized by the inclination $\alpha$, height $h$, and length $\ell$. 
$\alpha$ is the angle of the terrain with respect to the horizontal plane. 
$h$ is the vertical distance from the endpoint of the upper terrain to the lower terrain, and its sign implies an increase or decrease in elevation. The height of the last terrain is set to 0. $\ell$ is the line segment length.

\subsubsection{Simplified Robot Model}
\label{subsubsec:Robot Plannar Model}
For robots with flippers of different gear diameters, the skeleton model proposed in \cite{yuan2019configuration} cannot accurately capture the contact situation. 
To overcome this limitation, we introduce a planar model that covers the bottom of the robot with three rods of time-varying length, as shown in Fig. \ref{fig:planar_model}. 
We use the shorthand notation of $\left( \cdot \right)_{f,r}$ to denote the parameters of the front and rear flippers, respectively.
The angle of the model flippers is $\tilde{\theta}_{f,r}=\theta_{f,r}+\delta$, where $\delta=\arcsin\left((R-r)/l_0\right)$ is the angle caused by the gear diameter difference. 
This simplification approximates the arc at the large gear as a fold, and the error can be neglected when the flipper angle is small.
Furthermore, we extended the flipper length for capturing the outer end of the flippers in contact with the terrain, as depicted in Fig. \ref{fig:contact_approx}.
Specifically, we have $\tilde{l}_{f,r}=l_{f,r}\cos \delta +l_{f,r}^{i}+l_{f,r}^{o}$ and $\tilde{l}_t=l_t+l_{f}^{i}+l_{r}^{i}$, 
where
\vspace{-0.1cm}
\begin{equation}
\begin{cases}
	l_{f,r}^{i}\!=\!R\tan \left( \tilde{\theta}_{f,r}/2 \right)\\
	l_{f,r}^{o}\!=\!\max \left\{ r\tan \left( \left( \tilde{\theta}_{f,r}\mp \left( \theta _t+\alpha _{f,r} \right) \right) \!\!/2 \right) ,0 \right\}\\
\end{cases}\!. 
\end{equation}
For simplicity, we use symbols without tilde to represent the parameters of the robot model in the subsequent sections.

\setstretch{1.0}

\subsubsection{Configuration Space Reduction}
\label{subsubsec:Degrees of Freedom Reduction}
\begin{figure}[!t]
    \centering
    \subfloat[Definition of $s_d$ at ascent and descent.]{
        \centering
        \includegraphics[width = 7.8cm]{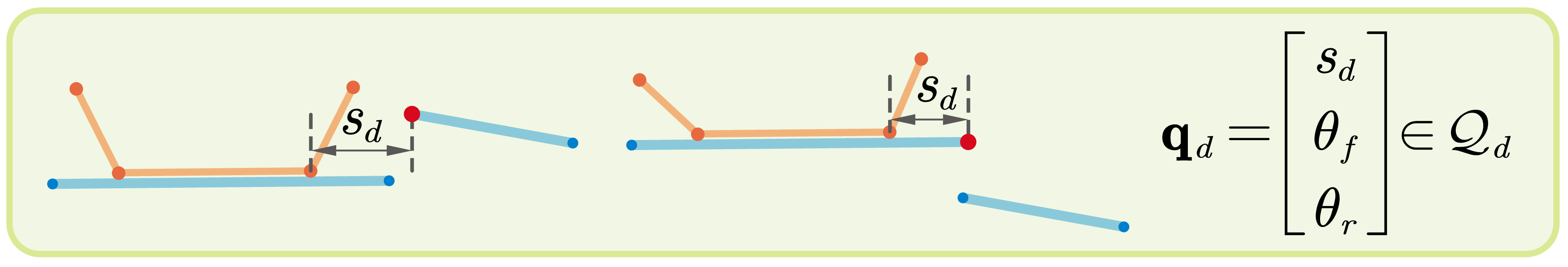}
        \label{fig:sd_def}
        \vspace{-0.4cm}
    }
    \vspace{-0.3cm}
    \vfill
    \subfloat[Definition of $s_t$ at ascent and descent.]{
        \centering
        \hspace{-0.25cm}
        \includegraphics[width = 7.8cm]{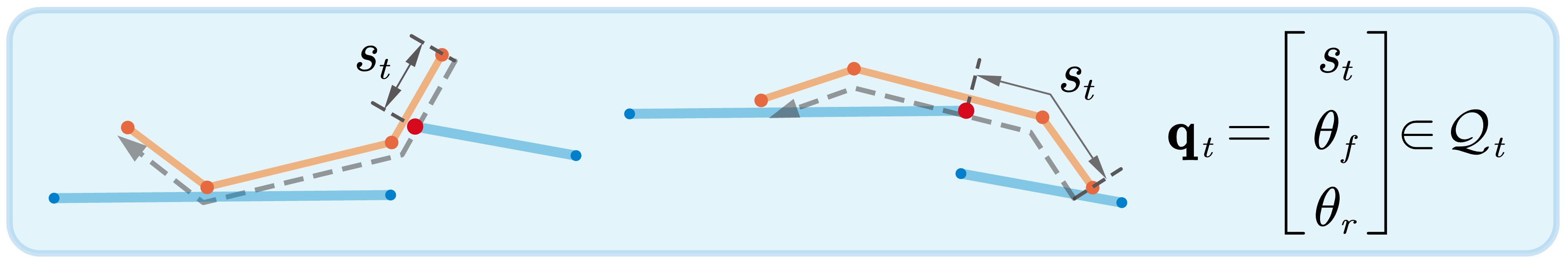}
        \label{fig:s_def}
    }
    \vspace{-0.15cm}
    \caption{
    Definitions of the generalized coordinates and configuration description, with the robot traveling from left to right. 
    Support points are marked with red dots. 
    The direction of increasing $s_t$ is indicated by dashed arrows. 
    }
    \vspace{-0.65cm}
    \label{fig:s_and_sd_definition}
\end{figure}

The robot state in the 2D plane can be expressed as
$
    \mathbf{\bar{q}}\!\!=\!\!\left[ \begin{matrix}
    x&	\!\!\!\!	z&	\!\!\!\!	\theta _t&	\!\!\!\!	\theta _f&	\!\!\!\!	\theta _r\\
    \end{matrix} \right] ^{\top}\!\!\in\!\! \mathbb{R} ^5,
    \label{equ:state_variables}
$
where $(x,z)$ denotes the position of the body center.  
During terrain traversal, there are at least two contact points between the robot and the terrain model, which reduces the effective DOFs.
We refer to the traversal as ascending when the robot traverses adjacent terrains with $h>0$, and as descending when $h<0$.
When ascending, the support point is defined as the starting point of the target terrain, and when descending, it is defined as the end point of the current terrain, which is illustrated in Fig. \ref{fig:s_and_sd_definition}. 
When the robot's track is attached to the ground, we define the distance from the front end of the track to the projection of the support point on the current terrain as $s_{d}$, then the robot state can be reduced to
$\mathbf{q}_d\!\!=\!\!\left[ \begin{matrix} s_d\!\!\!&	\theta _f\!\!\!&	\theta _r\\ \end{matrix} \right] \!^{\top}\!\in\!\mathcal{Q} _d\!\subset\! \mathbb{R} ^3$,
as shown in Fig. \ref{fig:sd_def}.
When the robot contacts the support point, we can describe its configuration by the generalized coordinate of the distance $s_t$ that the support point moves along the robot planar model, and the state is 
$\mathbf{q}_t\!\!=\!\!\left[ \begin{matrix} s_t\!\!\!\!&	\theta _f\!\!\!\!&	\theta _r\\ \end{matrix} \right]\! ^{\top}\!\in\mathcal{Q} _t\subset \mathbb{R} ^3$, as shown in Fig. \ref{fig:s_def}. 
$\mathcal{Q}_d$ and $\mathcal{Q}_t$ are driving and traversing configuration spaces, respectively, describing the model state in the two hybrid modes (see Sec. \ref{subsec:Traversal Sequence}).

\subsection{Hybrid Mode Definition and Traversal Sequence}
\label{subsec:Traversal Sequence}
\begin{figure}[!t]
    \centering
    \includegraphics[width = 0.79\linewidth]{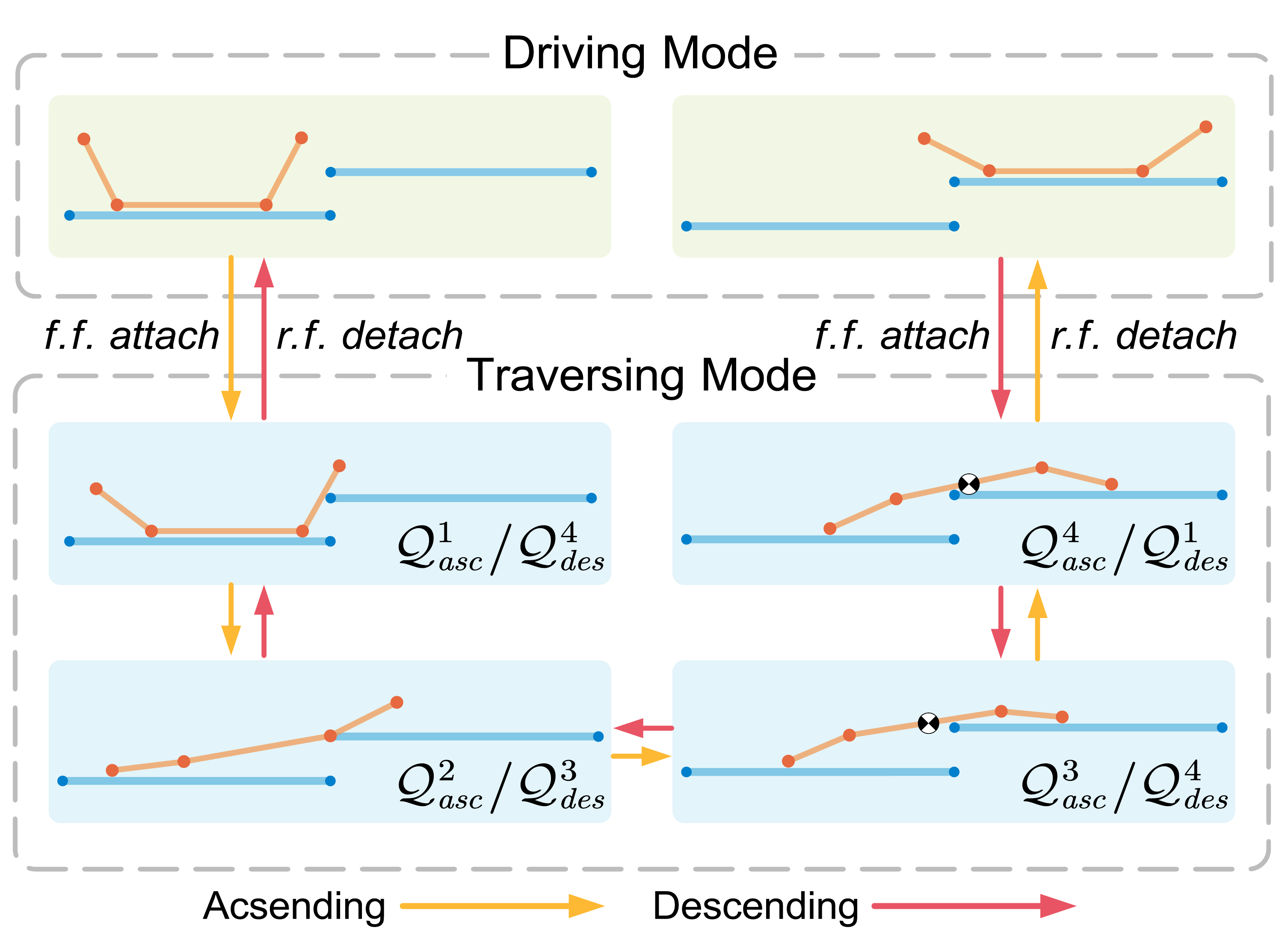}
    \caption{
    Illustration of our stable traversal strategy for ascent and descent, involving switching between driving mode and traversing mode.
    The lower four key nodes form the traversal sequence, counterclockwise for the ascending $S_{asc}$ and clockwise for the descending $S_{des}$.
    The abbreviations \textit{f.f.} and \textit{r.f.} are for the front flipper and rear flipper respectively.
    }
    \vspace{-0.75cm}
    \label{fig:traversal_sequence}
\end{figure}

In this part, we first introduce a stable traversal strategy, then define two hybrid modes based on the contact situation and propose a traversal sequence for HTO.
This strategy makes full use of the support of the flippers in traversal, as depicted in Fig. \ref{fig:traversal_sequence}. 
Specifically, 
during ascent, the front flipper spreads to attach to the support point and then climbs up. 
Then, the rear flipper supports the lower terrain while the front flipper supports the higher one, ensuring stability until the center of mass (COM) reaches the higher terrain. 
Once the COM reaches the higher terrain, the robot can lift the flippers and drive forward to complete the traversal.
During descent, the rear flipper supports to maintain a specific body inclination. This enables the front flipper to support the lower terrain, allowing the robot to drive and descend.

Based on the contact of the flippers, we can define two hybrid modes: \textit{driving} mode and \textit{traversing} mode. 
During the ascent, traversing mode starts when the front flipper attaches to the support point until the COM reaches higher terrain and then the rear flipper is able to detach from the lower terrain; during the descent,  it starts when the front flipper attaches the lower terrain and ends when the rear flipper detaches from the higher terrain.  
Additionally, the robot is considered to be in driving mode when it can drive without adjusting flippers.
When using the symbol $\mathbf{q}$, its meaning $\mathbf{q}_d$ (driving) or $\mathbf{q}_t$ (traversing) is determined by the current mode.

The concept of traversal sequence was first proposed in \cite{ohno2007semi}, but lacked a consistent formulation guideline. 
We propose a uniform guideline for this sequence to function similarly to the contact schedule of legged robots.
We define the traversal sequence as $ S=\left( \mathcal{Q}^{1},\cdots ,\mathcal{Q}^{N_s} \right) $, where $ \mathcal{Q}^{i}\subset \mathcal{Q} _t $ and $s_{t}^{i}\le s_{t}^{j} (i<j)$.
It consists of constrained configuration space key nodes that the robot must pass through during traversal. 
We proposed a traversal sequence with $N_s\!\!=\!\!4$, as shown in the four ordered blocks at the bottom of Fig. \ref{fig:traversal_sequence}.
Based on the symmetry of the robot configuration, the ascending sequence $S_{asc}=\left( \mathcal{Q} _{asc}^{1},\mathcal{Q} _{asc}^{2},\mathcal{Q} _{asc}^{3},\mathcal{Q} _{asc}^{4} \right)$ 
and the descending sequence $S_{des}=\left( \mathcal{Q} _{des}^{1},\mathcal{Q} _{des}^{2},\mathcal{Q} _{des}^{3},\mathcal{Q} _{des}^{4} \right)$ are naturally designed as symmetric processes, which is useful when formulating configuration descriptions.

The configuration descriptions for $\mathcal{Q}_{asc}^i$ can be 
geometrically derived by projecting the rod onto the vertical direction, 
as shown in Table \ref{tab:geo_cons}, where $s_t^{com}$ is the COM position expressed in generalized coordinate $s_t$.
To simplify the derivation, $\alpha$ and $h$ refer to the relative inclination and height between the two terrains, respectively, and $\theta_t$ is the pitch angle relative to the current terrain.
$\left( \cdot \right) _u$ and $\left( \cdot \right) _l$ are used to denote the upper and lower bounds.

By utilizing the symmetry of the robot configuration and the sequence, we do not need to re-derive the descending configuration description.
We define the dual traversing configuration as
$\mathbf{q}_{t}^{*}\!= \!\left[ \begin{matrix} l_{\Sigma}-s_t&	\!\!\theta _r&\!\!		\theta _f\\ \end{matrix} \right] ^{\top}$,
where $l_{\Sigma} = l_f+l_t+l_r$. 
This can be considered as the forward direction of the robot being flipped.
Additionally, the parameters in the ascending configuration $\mathcal{Q}_{asc}$ are replaced with their dual counterparts, which involves swapping $l_f$ and $l_r$, $\theta_f$ and $\theta_r$, and $s_t^{com}$ and $l_\Sigma-s_t^{com}$ to obtain the dual description $\mathcal{Q} _{asc}^* $.
As a result, the descending configuration $\mathbf{q}_t\in \mathcal{Q} _{des}^{i}$ can be described by $ \mathbf{q}_{t}^{*}\in \left( \mathcal{Q} _{asc}^{5-i} \right) ^* $.

Additionally, for the special case of $h=0$, we adopt the descending description when $\alpha<0$, as it is compatible with the descending strategy. For $\alpha>0$, we can introduce a height of $h_\delta>0$ to adopt the ascending description.

\newcommand{\ci}{Node}
\newcommand{\cii}{Configuration Description}
\newcommand{\ciii}{$\mathcal{Q}_{asc}^{1}$}
\newcommand{\civ}{$0<s_t<l_f$}
\newcommand{\cv}{$\begin{array}{l}
          \left( l_f-s_t \right) \sin \theta _f=h\\
          0\le \theta _r \le \theta_{u}\\
          \end{array}$}
\newcommand{\cvi}{$\mathcal{Q}_{asc}^{2}$}
\newcommand{\cvii}{$s_t=l_f$}
\newcommand{\cviii}{$
    \begin{array}{l}
    	l_t\sin \theta_t+l_r\sin \left( \theta_t-\theta_r \right) =h\\
    	\theta _{t,l}\le \theta_t\le \theta _{t,u}\\
    	\theta_t-\alpha \le \theta_f\le \theta_{u} \\
    	\theta_l \le \theta_r\le \theta_t\\
    \end{array}
    $}
\newcommand{\cix}{$
\begin{array}{c}
	\\
	\mathcal{Q} _{asc}^{3}\\
	\\
\end{array}
$}    
\newcommand{\cx}{$l_f<s_t<s_t^{com}$}    
\newcommand{\cxi}{$
    \begin{array}{l}
        \left( s_t-l_f \right) \sin ( \theta_t-\alpha )\\
        +l_r\sin ( \theta_t+\theta _f-\alpha ) =0\\
        \left( l_t+l_f-s_t \right) \sin \theta_t\\
        +l_r\sin ( \theta_t-\theta _r ) =h\\
        \alpha \le \theta_t\le \theta _{t,u},
        \theta_l \le \theta_r\le \theta_t\\
    \end{array}
$}    
\newcommand{\cxii}{
\setlength{\extrarowheight}{0.05cm}
$
\begin{array}{c}
	\\
	\mathcal{Q} _{asc}^{4}\\
	\\
\end{array}
$
\setlength{\extrarowheight}{-0.0cm}
}    
\newcommand{\cxiii}{$s_t^{com}\leq s_t<l_f+l_t$}    
\begin{table}[!htbp]
    \centering
    \renewcommand\arraystretch{1.15}
    \caption{Configuration Descriptions On Key Nodes During Ascending}
    \resizebox{1.0\linewidth}{!}{ 
    \begin{tabular}{|c|cl|}
    \hline
    \ci & \multicolumn{2}{c|}{\cii} \\ \hline
    \ciii & \multicolumn{1}{c|}{\civ}    & \cv   \\ \hline
    \cvi  & \multicolumn{1}{c|}{\cvii}   & \cviii  \\ \hline
    \cix  & \multicolumn{1}{c|}{\cx}     & \multirow{2}{*}{\cxi}   \\ \cline{1-2}
    \cxii & \multicolumn{1}{c|}{\cxiii}  &             \\ \hline
    \end{tabular}
    }
    \label{tab:geo_cons}
    \vspace{-0.35cm}
\end{table}
\section{Hybrid Trajectory Optimization}
\label{sec:methods}
\setstretch{0.97}
\subsection{Problem Formulation}
\label{subsec:traj_opt_prob}
We formulate the HTO problem for terrain traversal with the multi-objective cost function ((\ref{equ:cost}), see Sec. \ref{subsec:Cost Function})) weighted by $w_{i}\left(i=1,2,3\right)$, and constraints ((\ref{equ:initial})-(\ref{equ:switch_constr}), see Sec. \ref{subsec:Constraints}). 
The objective is to obtain the trajectory $\left\{s_d\!\left( t \right), s_t\!\left( t \right), \theta _f\!\left( t \right), \theta _r\!\left( t \right)\right\}$. 
The driving and traversing intervals are denoted as $\mathcal{T}_d$ and $\mathcal{T}_t$, respectively, with $t_{S}^{\left(k\right)} \left(k=1\cdots,n_S\right)$ and $t_T$ representing the mode switching moments and the end of the planning horizon respectively.
\vspace{-0.1cm}
\begin{subequations}
    \begin{align}
    \!\!\min\quad&J =w_1 J_{time}+ w_2 J_{coh}+ w_3 J_{stab} \label{equ:cost}
    \\
    \mathrm{s.t.}\quad 
    &\mathrm{initial\;state:\;} \mathbf{q}\left( 0 \right) =\mathbf{q}_0, \dot{\mathbf{q}}\left( 0 \right) =\dot{\mathbf{q}}_0 \label{equ:initial}
    \\
    &\mathrm{terminal\;state:\;} \mathbf{q}\left( t_T \right) = \mathbf{q}_T \label{equ:terminal}
    \\
    &\mathrm{configuration}
    \begin{cases}
    	\mathbf{q}\left( t\in \mathcal{T} _d \right) \in \mathcal{Q} _d\\
    	\mathbf{q}\left( t\in \mathcal{T} _t \right) \in \mathcal{Q} _{asc},\enspace \mathrm{for}\;\mathrm{asc.}\\
    	\mathbf{q}^*\left( t\in \mathcal{T} _t \right) \in \mathcal{Q} _{asc}^{*},\mathrm{for}\;\mathrm{des.}\\
    \end{cases}
    \label{equ:config_constr}
    \\
    &\mathrm{motion:\;}\mathbf{\dot{q}}\left( t \right) =\mathbf{u}\in \mathcal{U}  \label{equ:motion_constr}
    \\
    &\mathrm{switching:\;}
    \begin{cases}
    	\Delta _c\left( \mathbf{q}^-( t_{S}^{\left(k\right)} ) ,\mathbf{q}^+( t_{S}^{\left(k\right)} ) \right) =0\\
    	\Delta _m\left( \mathbf{q}^-( t_{S}^{\left(k\right)} ) ,\mathbf{q}^+( t_{S}^{\left(k\right)} ) \right) \le 0\\
    \end{cases}
    \label{equ:switch_constr}
    \end{align}
    \label{equ:to_problem}
    \vspace{-0.1cm}
\end{subequations}
\vspace{-0.2cm}

Compared to the shooting approach \cite{chen2022quadruped} with fixed time steps to plan the hybrid motion, we adopt the direct collocation method \cite{kelly2017introduction} to transcribe \ref{equ:to_problem} into a nonlinear programming (NLP) problem and plan the durations between nodes, which has a higher planning flexibility.
Since the traversal sequence is predefined and the hybrid mode is time-dependent switching, the decision variables are set to the node state $\left( \mathbf{q}, \mathbf{\dot{q}}\right)$, as well as the durations $T_{d,i}^{\left(k\right)} (i\!=\!1,\cdots,n_d^{\left(k\right)}-1)$, $T_{t,j}^{\left(k\right)} (j\!=\!1,\cdots,n_t^{\left(k\right)}-1)$.
The continuous time trajectory is a smooth piecewise cubic polynomial curve, generated by Hermite interpolation.

\subsection{Receding-Horizon Planning}
\label{subsec:Receding Horizon Planning Schedule}
\begin{figure}[!t]
    \centering
    \subfloat[Planning horizon in driving mode.]{
        \includegraphics[width = 0.85\linewidth]{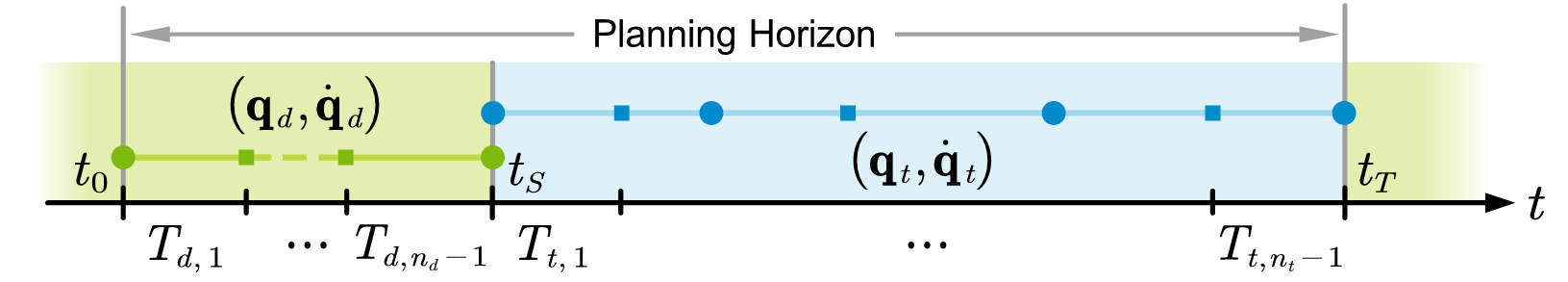}
        \label{fig:planning_sequence_a}
    }
    \vfill
    \vspace{-0.2cm}
    \subfloat[Planning horizon in traversing mode. ]{
        \includegraphics[width = 0.85\linewidth]{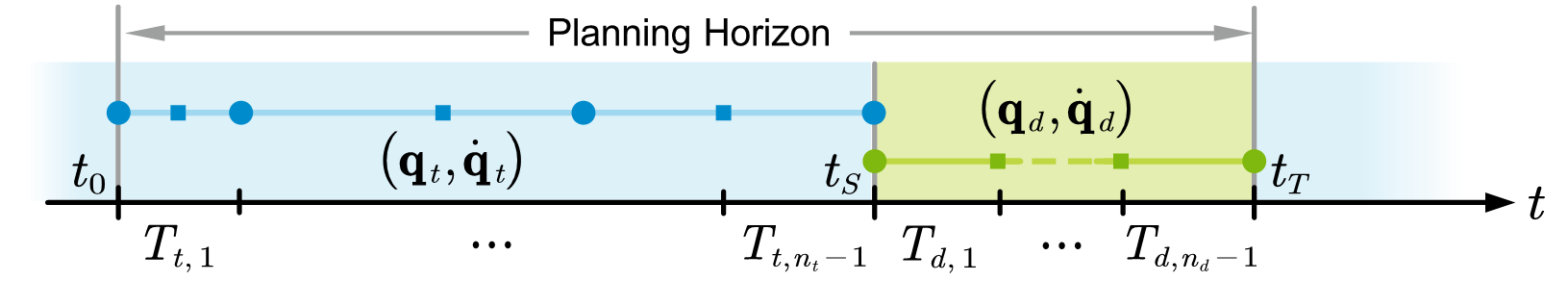}
        \label{fig:planning_sequence_b}
    }
    \vfill
    \vspace{-0.25cm}
    \subfloat{
        \includegraphics[width = 0.8\linewidth]{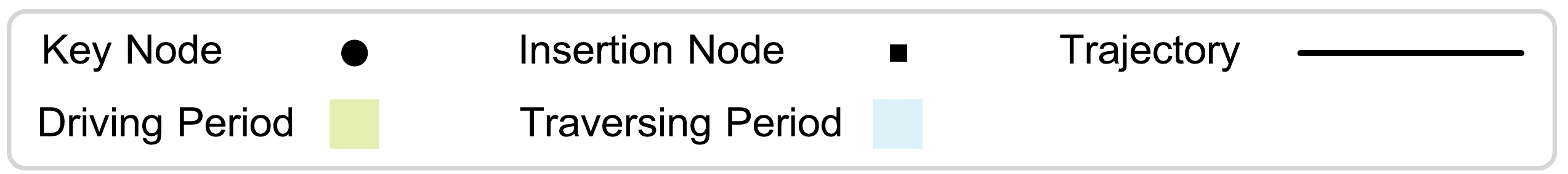}
        \label{fig:planning_sequence_legend}
    }
    \vspace{-0.1cm}
    \caption{
    Planning horizons for the subproblem that contains a single mode switch.
    Key nodes in the traversal sequence are represented by circular dots. Meanwhile, additional insertion nodes can be added, depicted by square dots, and have geometrically derived configuration constraints.
    Insertion nodes influence the trajectory trend and affect the scale of the subproblem.
    }
    \label{fig:planning_sequence}
    \vspace{-0.5cm}
\end{figure}

The HTO problem (\ref{equ:to_problem}) describes the complete task of multi-terrain traversal with possible multiple mode switches, including a large number of nodes and constraints. In addition, the configuration constraints involve state-dependent time-varying parameters $(l_f, l_t, l_r)$, rendering real-time trajectory optimization arduous and posing challenges to trajectory robustness and adaptability.
To tackle this problem and facilitate real-time planning, we divide the task into a series of subproblems that contain only one mode switch, \textit{i.e.}, $n_S\!\!=\!\!1$, either from driving to traversing (Fig. \ref{fig:planning_sequence_a}) or from traversing to driving (Fig. \ref{fig:planning_sequence_b}), 
and the time-varying parameters adopt fixed values at the time of problem update. 
The planning horizon $t_T$ is the moment of the next mode switch, guaranteed by configuration constraints.
By employing receding-horizon planning, the subproblem is updated at a fixed frequency for continuous replanning to ensure kinematic feasibility and enhance robustness.

\subsection{Constraints}
\label{subsec:Constraints}
To ensure that the robot follows the predefined traversal sequence, we impose configuration constraints at each node. 
The motion feasibility is guaranteed by the motion constraints. 
Moreover, we enforce switching constraints for mode transitions.

\subsubsection{Configuration Constraints}
At the key nodes of the driving mode, we enforce the track to be fit to the ground, and the flippers are constrained as
$
    \mathcal{Q} _d=\left\{ \mathbf{q}_d\mid 0\le \theta _{f,r}\le \theta_{u} \right\}. 
$
The traversing mode configuration constraints are given in Table \ref{tab:geo_cons}, with the pitch angle $\theta_t$ introduced as an auxiliary variable. 
For the descent, we impose the dual constraints on its dual configuration $\mathbf{q}^*$.

\subsubsection{Motion Constraints}
\label{subsubsec:Motion Constraints}
To ensure the motion feasibility, we apply (\ref{equ:motion_constr}) to bound the nominal input $\mathbf{u} = \mathbf{\dot{q}}(t)$,
where $\mathcal{U}=\left\{\mathbf{u}\mid 0\leq v\leq v_{u},\omega_{l}\leq\omega_{f,r}\leq\omega_{u}\right\}$ and $\omega_{f,r}$ denotes the flipper angular velocities. The nominal velocity is defined as:
\vspace{-0.3cm}
\begin{align}
    v=\begin{cases}
    	-\dot{s}_d,&		\mathrm{if}\;\mathbf{q}\in \mathcal{Q} _d\\
    	\dot{s_t},&		\mathrm{if}\;\mathbf{q}\in \mathcal{Q} _t\\
    \end{cases}.
    \label{equ:velocity}
\end{align}
\vspace{-0.2cm}
\subsubsection{Switching Constraints}
\label{subsubsec:Mode Switching Constraints}
We associate the predefined traversal sequence with durations to describe the mode switching events, and determine the switching moment $t_S$ by optimizing durations under the switching constraints.
At $t=t_S$, we impose configuration and motion constraints to achieve mode transitions between pre- and post-switch states $\mathbf{q}^{-}$ and $\mathbf{q}^{+}$.
By projecting $s_t$ onto the direction of $s_d$, the configuration switch constraint function from driving to traversing is formulated as:
\begin{equation}
    \!\!\!\!\Delta_c ( \mathbf{q}^-\!\!\!,\mathbf{q}^+\! ) \!\!=\!\!\begin{cases}
     \!\left( l_f\!-\!s_t \right)\! \cos \theta _f\!-\!s_d
     , \mathrm{if}\;\mathbf{q}^-\!\!\!\in\! \mathcal{Q}_d,  \mathbf{q}^+\!\!\in\! \mathcal{Q}_{asc}^{1}\!\!\!\!\\
    \!\left( l_f\!-\!s_t \right)\! \cos \theta _t\!-\!s_d, \mathrm{if}\;\mathbf{q}^-\!\!\!\in\! \mathcal{Q}_d,  \mathbf{q}^+\!\!\in\! \mathcal{Q}_{des}^{1}\!\!\!\!\\
    \end{cases}\!\!. 
    \label{equ:switch_constr_config_def}
\end{equation}

The motion-switching constraint considers the velocity loss.
By projecting velocity into the $s_t$ direction, we formulate this function from driving to traversing as follows: 
\vspace{-0.2cm}
\begin{equation}
    \!\!\!\Delta _m( \mathbf{q}^-\!\!\!,\mathbf{q}^+\! ) \!\!=
    \!\!\begin{cases}
    	\!\dot{s}_t+\dot{s}_d\cos \theta _f, \mathrm{if}\;\mathbf{q}^-\!\in \mathcal{Q} _d, \mathbf{q}^+\!\in \mathcal{Q} _{asc}^{1}\\
    	\!\dot{s}_t+\dot{s}_d\cos \theta _t, \mathrm{if}\;\mathbf{q}^-\!\in \mathcal{Q} _d, \mathbf{q}^+\!\in\mathcal{Q} _{des}^{1} \!\!\!\!\\
    \end{cases}\!\!\!\!. 
    \label{equ:switch_constr_motion_def}
\end{equation}
The functions for traversing to driving are also formulated through a similar derivation.

\subsection{Cost Function}
\label{subsec:Cost Function}
The multi-objective cost function combines time, coherence, and stability costs, to enhance the traversal quality.
\setstretch{1.0}
\subsubsection{Time Cost}
To minimize the time required for traversal, we formulate the time cost as follows:
\vspace{-0.15cm}
\begin{equation}
J_{time}=\sum_{i=1}^{n_d-1}{T_{d,i}^{2}}+\sum_{i=1}^{n_t-1}{T_{t,i}^{2}}. 
\end{equation}
\vspace{-0.2cm}

\subsubsection{Coherence Cost}
To enhance motion coherence, this cost mitigates the unnecessary back-and-forth motion caused by polynomial interpolation by penalizing the difference between the average rate of change between adjacent nodes and the derivatives, thereby improving energy efficiency. We formulate it as: 
\vspace{-0.1cm}
\begin{equation}
J_{coh}=\sum_{\mathbf{q}_i\in \mathcal{Q} _d, \mathcal{Q} _t}^{}{\left( \left\| \mathbf{\bar{\dot{q}}}_i-\mathbf{\dot{q}}_i \right\| _{\mathbf{Q}_s}^{2}+\left\| \mathbf{\bar{\dot{q}}}_i-\mathbf{\dot{q}}_{i+1} \right\| _{\mathbf{Q}_s}^{2} \right)}, 
\vspace{-0.1cm}
\end{equation}
where 
$
\left\| \mathbf{x} \right\| _{\mathbf{Q}}\coloneqq \sqrt{\mathbf{x}^{\top}\mathbf{Qx}}
$, $\mathbf{\bar{\dot{q}}}_i\!\coloneqq\!\left( \mathbf{q}_{i+1}-\mathbf{q}_i \right) /T_i$ and $\mathbf{Q}_s$ is a positive-definite matrix.

\subsubsection{Stability Cost}
The stability cost, inspired by the strategy selector in \cite{gianni2016adaptive}, enables the robot to adjust its flipper configurations based on terrain conditions to maintain stability. 
In sparse terrains such as stairs, the flippers tend to flatten to increase the contact area and lower the COM to improve stability; whereas in dense terrains, the flippers can lift to balance motor workload.
This cost penalizes the deviations between the angles of the flippers and the terrain on which they are located in sparse terrains, aiming to lower the COM and increase the contact area.
We formulate it as: 
\begin{equation}
    \!\!\!\!J_{stab}\!=\!\!\!\!\!\sum_{i=1}^{n_d+n_t-1}{\!\!\!\!\left( \!w_{f,i}\left(\theta _{t,i}\!\!+\!\theta _{f,i}\!-\!\alpha _{f,i} \right) ^2\!\!+\! w_{r,i}\left(\theta _{t,i}\!-\!\theta _{r,i} \right)^{2} \right)},\!
    \vspace{-0.1cm}
\end{equation}
where $\alpha_{f}$ takes 0 in driving mode and the relative angle $\alpha$ in traversing mode. 
The weights $w_f$ and $w_r$ depend on the terrain sparsity, which will be defined in Sec. \ref{subsec:Terrain_Simplification}. 
Considering the sparsity of the current terrain $c(\tau_1)$ and the target terrain $c(\tau_2)$, we have $w_r=c(\tau_1)$ and 
$w_f$ as follows:
\vspace{-0.25cm}
\begin{align}
    \!\!w_f=\begin{cases}
    c(\tau_1),&	\!\!\mathrm{if}\;\mathbf{q}\in \mathcal{Q} _d\\
    0,&	\!\!\mathrm{if}\;\mathbf{q}\in \mathcal{Q} _{asc}^{1}\,\,\mathrm{or}\,\,\, \mathbf{q}\in \mathcal{Q} _{des}^{4}\\
    c(\tau_2),&	\!\!\mathrm{if}\;\mathbf{q}\in \mathcal{Q} _{asc}\backslash \mathcal{Q} _{asc}^{1}\,\,\mathrm{or}\,\,\, \mathbf{q}\in \mathcal{Q} _{des}\backslash \mathcal{Q} _{des}^{4}\\
    \end{cases}.
\end{align}
\vspace{-0.15cm}

\section{Terrain Traversal System}
\label{sec:framework}

\subsection{Overview}
\label{subsec:overview}

\begin{figure}[!t]
    \centering
    \includegraphics[width = 0.95\linewidth]{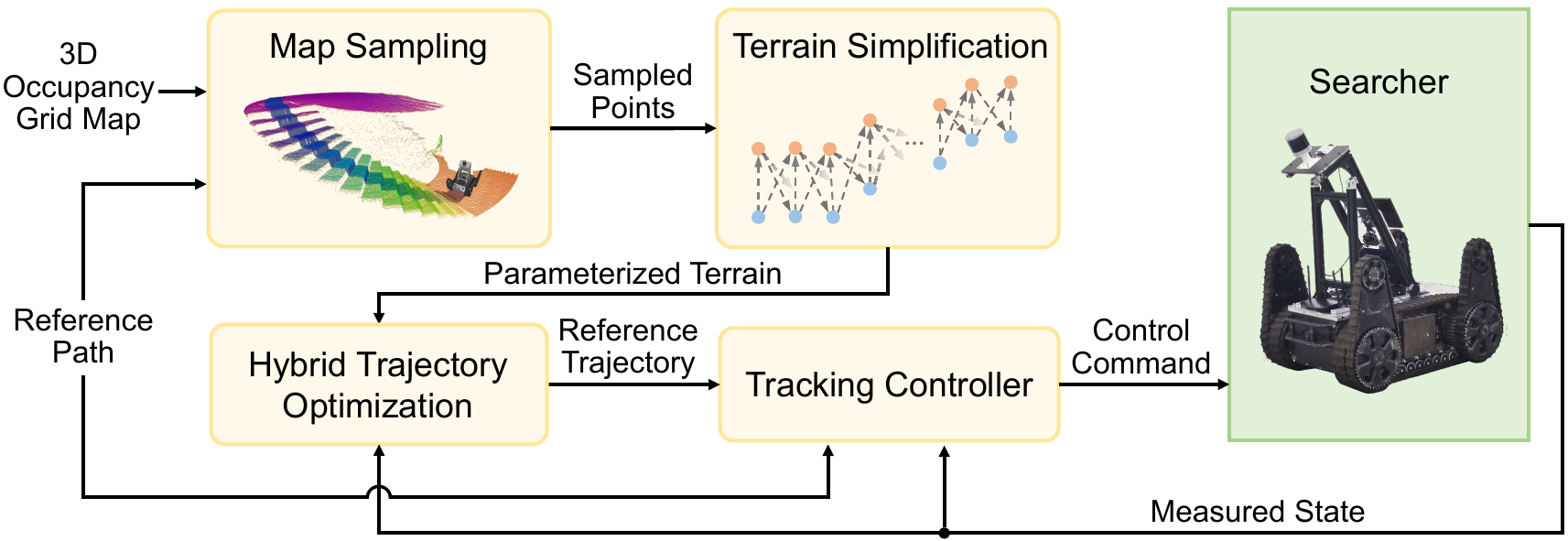}
    \caption{
    Workflow of the terrain traversal system. 
    The map sampling module operates on a 3D occupancy grid map and samples each waypoint along the normal of the reference path to obtain a sampled strip.
    Subsequently, the sampled strip is filtered on the normal to mitigate the effects of map sparsity and random noise.
    Afterward, the mean height of each normal is projected onto the distance-height plane, generating a point set $P$, which is then passed to the terrain simplification module (Sec. \ref{subsec:Terrain_Simplification}). 
    The simplified terrain and measured robot states are sent to the HTO module.
    Finally, the tracking controller (Sec. \ref{subsec:Controller}) utilizes the reference path and trajectory to generate control commands.
    }
    \vspace{-0.6cm}
    \label{fig:workflow}
\end{figure}

Our proposed terrain traversal system comprises four modules: map sampling, terrain simplification, hybrid trajectory optimization, and tracking controller. Fig. \ref{fig:workflow} provides a schematic overview of the workflow for terrain traversal.

\subsection{Terrain Simplification}
\label{subsec:Terrain_Simplification}
\vspace{-0.1cm}
To obtain simplified terrains, we use a set of line segments $\mathbb{T}$ to cover a set of sampled points $P=\left\{p_i=\left( d_i,h_i \right)\right\}_{i=1}^{N}$ and transform the cover problem into an optimal path problem of graph theory.

The edge of the terrain provides vital support to the robot during traversal, so large errors in the edge points of the simplified terrain are unacceptable. 
Therefore, we choose the points in $P$ as the start and end points of the line segments, denoted as $\tau_{s,e}=(p_{s},p_{e}),d_s<d_e$. 
The line segment needs to cover the points in its range, \textit{i.e.}, for any $p_i \in P$ and $d_s\le d_i\le d_e$, there is $dist(p_i,\tau_{s,e})=\frac{p_{s}p_{e}\times p_{s}p_{i}}{\left\| p_{s}p_{e} \right\|}\le \delta_m$, where $\delta_m$ is the maximum height difference that the robot can tolerate in driving mode.
$p_i$ is considered to be within $\tau_{s,e}$ when $\left| dist(p_i,\tau_{s,e}) \right| \!\le\! \delta_{l}$, where $\delta_{l}$ represents tolerance.
To prevent the omission of large potholes, the distance between adjacent points within $\tau_{s,e}$ should not exceed ${l_t}/{2}$. Terrain sparsity is defined as 
\vspace{-0.1cm}
\begin{equation}
c(\tau_{s,e})=1-\frac{P\!N(\tau_{s,e})}{(d_e-d_s)/d_r+1}\in [0,1],
\end{equation}
where $\!P\!N(\tau)\!$ is the number of points within $\!\tau\!$ and $\!d_r\!$ is the sampling resolution.
Greater sparsity means greater morphological differences between the simplified and actual terrain.

The set of all simplified terrains that the robot can traverse is defined as $\mathbb{T}^* = \left\{ \tau_{s,e}^*\right\}$. To find the optimal coverage $\mathbb{T}\subset\mathbb{T}^*$, covering the sampled points using the fewest terrains while keeping as many points within the terrains as possible, we construct a weighted Directed Acyclic Graph (DAG) $G(V,E)$ with $2N$ vertices. $V=\left\{v_{2i-1},v_{2i} \right\} _{i=1}^{N}$, where each vertex $v_{2i-1}$ and $v_{2i}$ corresponds to using $p_i$ as the start and end points of a line segment.
The edge set $E$ should ensure a one-to-one correspondence between paths from $v_1$ to $v_{2N}$ and subsets of $\mathbb{T}^*$. The path's cost is the count of terrains in its corresponding subset, while the reward is the number of sampled points within all subset's terrains.
We use
$E_{s\rightarrow e}\!=\!\left\{ \left( v_{2s-1}\rightarrow v_{2e} \right) \mid \forall \tau _{s,e}^* \in \mathbb{T}^* \right\}$
to denote simplified line segment candidates, which have an edge cost of 1 and reward of $PN(\tau_{s,e}^*)$.
To ensure connectivity, we add auxiliary edges $E_{e\rightarrow s}\!=\!\left\{ \left( v_{2i},v_{2\left( i+j \right) -1} \right) \mid 1\le i\le N-1,0\le j\le N_{ign} \right\}$, where $N_{ign}$ denotes the number of ignored points. 
The cost of these edges is 0, as they do not represent line segments but merely links between the endpoint of the previous line segment and the start point of the next one.
When $j=0$, two adjacent line segments are directly connected, and the edge rewards are -1 to avoid double counting of the same point. 
When $j>0$, the distance between the two line segments does not exceed $N_{ign} \cdot d_r$. This approach enables the algorithm to ignore noise that may occur between adjacent planes with large height differences on the map, thus improving its robustness. The edge reward for this kind of edge is 0.
Finally, for a DAG with non-negative edge costs, we can obtain the minimum cost path from $v_{1}$ to $v_{2N}$ using Dijkstra's algorithm. Subsequently, we use a memoization search to efficiently acquire the optimal path with minimum cost and maximum reward. 

\vspace{-0.1cm}
\subsection{Tracking Controller}
\label{subsec:Controller}
\setstretch{0.965}
The tracking controller is designed to follow both the trajectory generated by HTO and the reference path.
The flipper velocity commands are calculated from the reference angles (the model flipper angles are converted to real angles) using the PD control law 
$\omega _{f,r}=K_pe_{\theta _{f,r}}+K_de_{\omega _{f,r}}$,
where $e_{\left( \cdot \right)}$ denotes the deviation between desired and measured.

The nominal velocity input can be computed using the trajectory of generalized coordinates, based on (\ref{equ:velocity}). 
It is essential to ensure that the robot's yaw angle aligns with the reference path while following the nominal velocity.
To achieve this, we calculate the robot's angular velocity $\omega$ by using (\ref{equ:omega_control_law}). 
The velocity input (\ref{equ:vel_control_law}) is a combination of the nominal velocity feed-forward $v^{des}$ and position feedback. 
The deviation $e_s$ from the generalized coordinates while driving needs to be taken as the opposite, since $s_d$ decreases.
\begin{align}
    \vspace{-0.6cm}
    &\omega =K_{p}^{\phi}e_{\phi}+K_{d}^{\phi}\dot{e}_{\phi},
    \label{equ:omega_control_law}
    \\
    &v=v^{des}+K_{p}^{s}e_s+K_{i}^{s}\int_0^t{e_s\left( \tau \right) \mathrm{d}\tau}. 
    \label{equ:vel_control_law}
    \vspace{-0.6cm}
\end{align}
Finally, $\left( v,\omega \right)$ is converted to the differential input $\left( v_{\mathrm{l}},v_{\mathrm{r}} \right)$ for the tracked robot, and the derivation can be found in \cite{gianni2016adaptive}.

\setstretch{1.0}
\section{Experiments and Results}
\label{sec:experimencts_and_results}
\begin{figure}[!tbp]
    \centering
    \includegraphics[width = 0.8\linewidth]{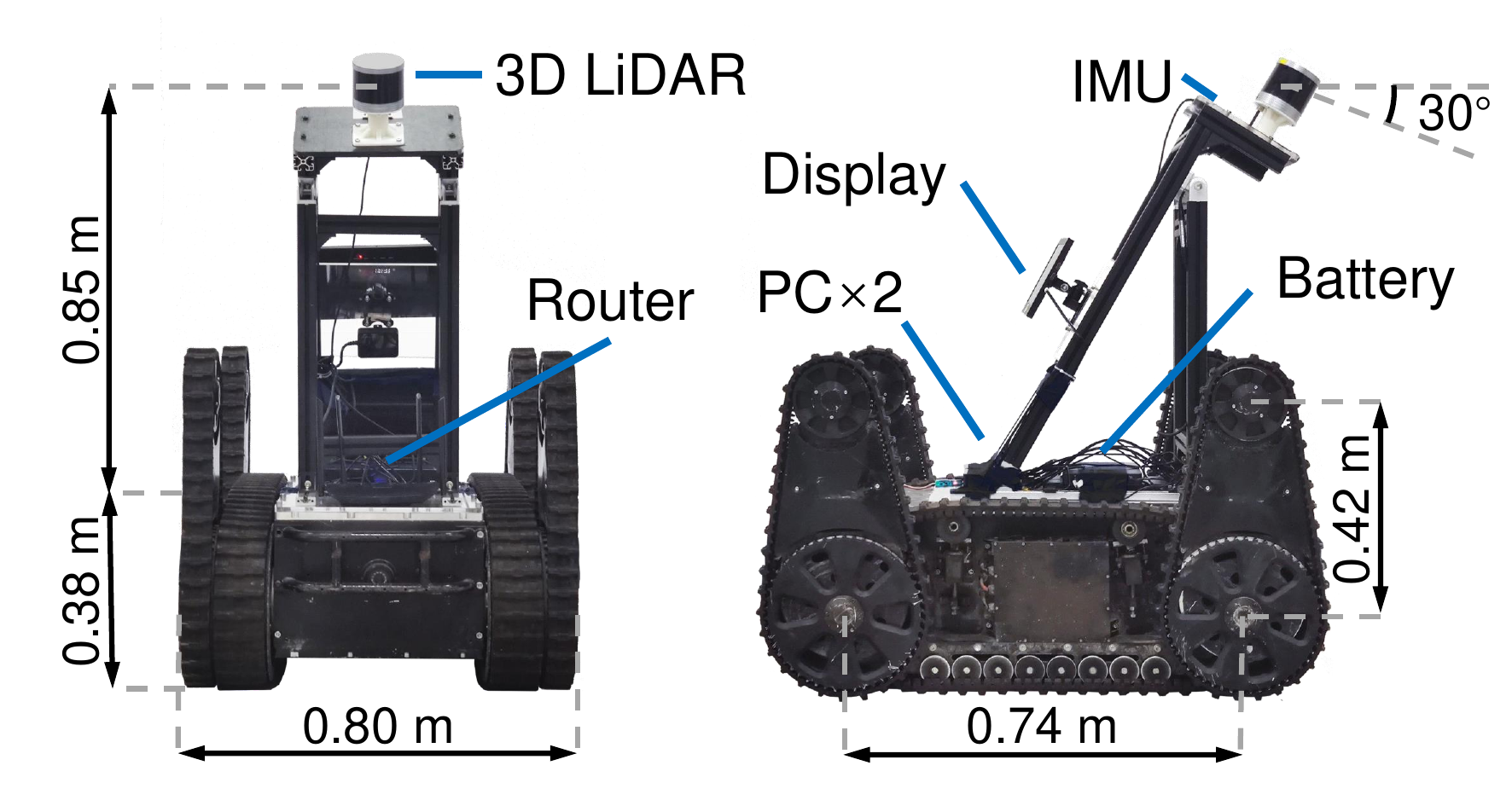}
    \vspace{-0.2cm}
    \caption{
    The Searcher is equipped with the RS-Helios 5515 32-beam LiDAR and IMU. 
    Two Intel NUC 8 (i7-8559U CPU@2.70GHz, 32GB RAM) are used, one for state estimation and the other for planning and control.
    }
    \label{fig:hardware}
    \vspace{-0.6cm}
\end{figure}

We conduct experiments in both simulation and real-world scenarios to verify the effectiveness of our work. 
The NLP problem for HTO is formulated in CasADi \cite{andersson2019casadi} and solved using IPOPT \cite{biegler2009large}. 
A priori 3D occupancy grid map with a resolution of 0.02m and a predefined reference path are used in all experiments.
We utilize FAST-LIO2\cite {xu2022fast} for localization.
The optimization nodes number is set to $n_t\!=\!5$ and $n_d$ is determined by the terrain length divided by 0.5m, which facilitates real-time planning (typical computation times are within 150ms, and the problem is updated at a fixed frequency of 5Hz).
We set the weights in all experiments to $ \left( w_1,w_2,w_3 \right) \!=\!\left( 1,0.3,350 \right)$ and $\mathbf{Q}_s=\mathrm{diag}\left( 1,0.8,0.8 \right)$. 
The reference trajectory is interpolated to 100Hz for execution. 
The fast simulation \cite{pecka2017fast} allows us to conduct real-time simulations of Searcher in Gazebo.
Fig. \ref{fig:hardware} shows the Searcher platform for real-world experiments. 
Demonstrations of all the experiments are available at \url{https://youtu.be/fL8w1WPwDAU}.

\vspace{-0.2cm}
\subsection{Traversal Quality Comparison in Typical Terrains}
\label{subsec:Simulation Experiments}
\begin{figure}[!t]
    \vspace{-0.2cm}
    \centering
    \subfloat[]{
    \centering
    \includegraphics[height = 0.385\linewidth]{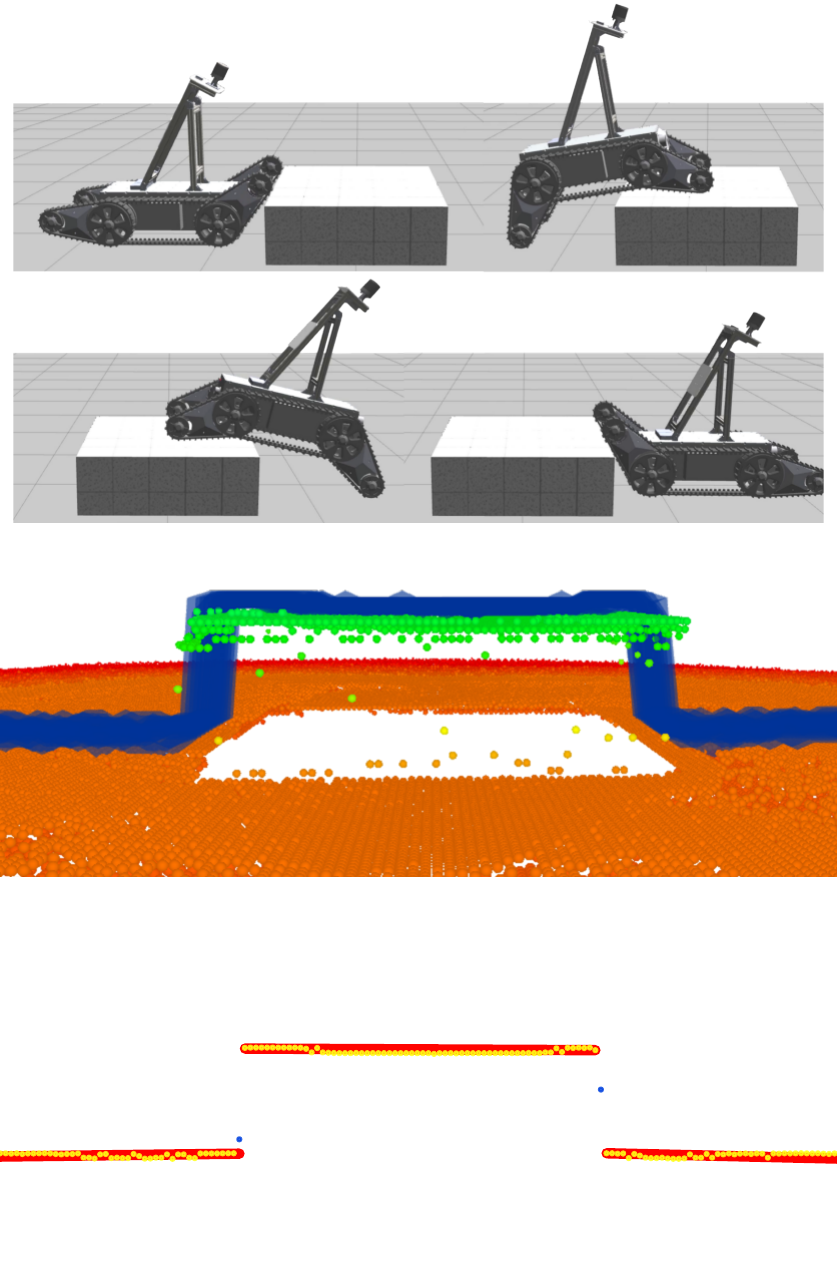}
    \label{fig:platform}
    }
    \hfill
    \subfloat[]{
    \centering
        \includegraphics[height = 0.37\linewidth]{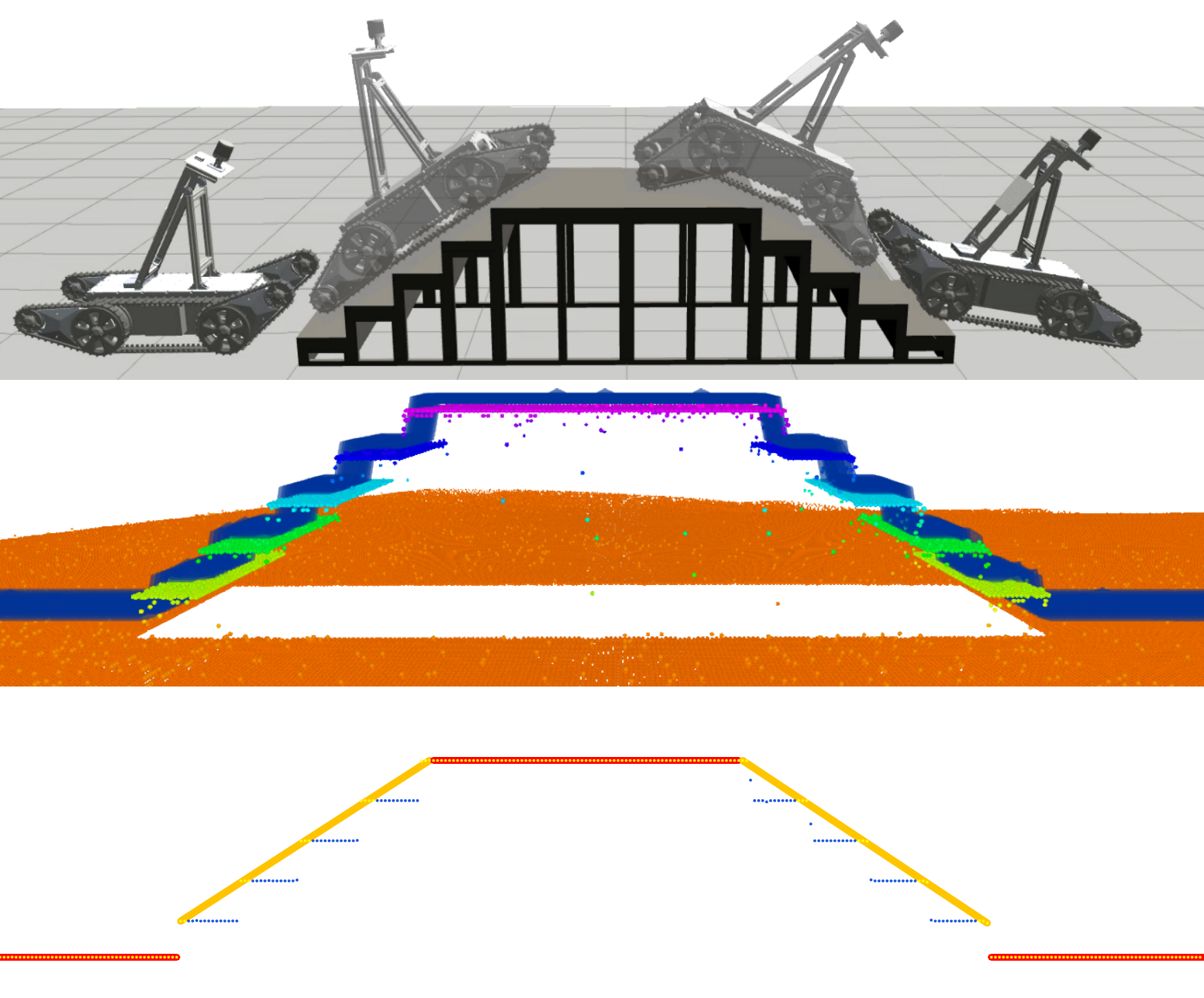}
        \label{fig:stairs}
    }
    \hfill
    \subfloat[]{
    \centering
    \includegraphics[height = 0.38\linewidth]{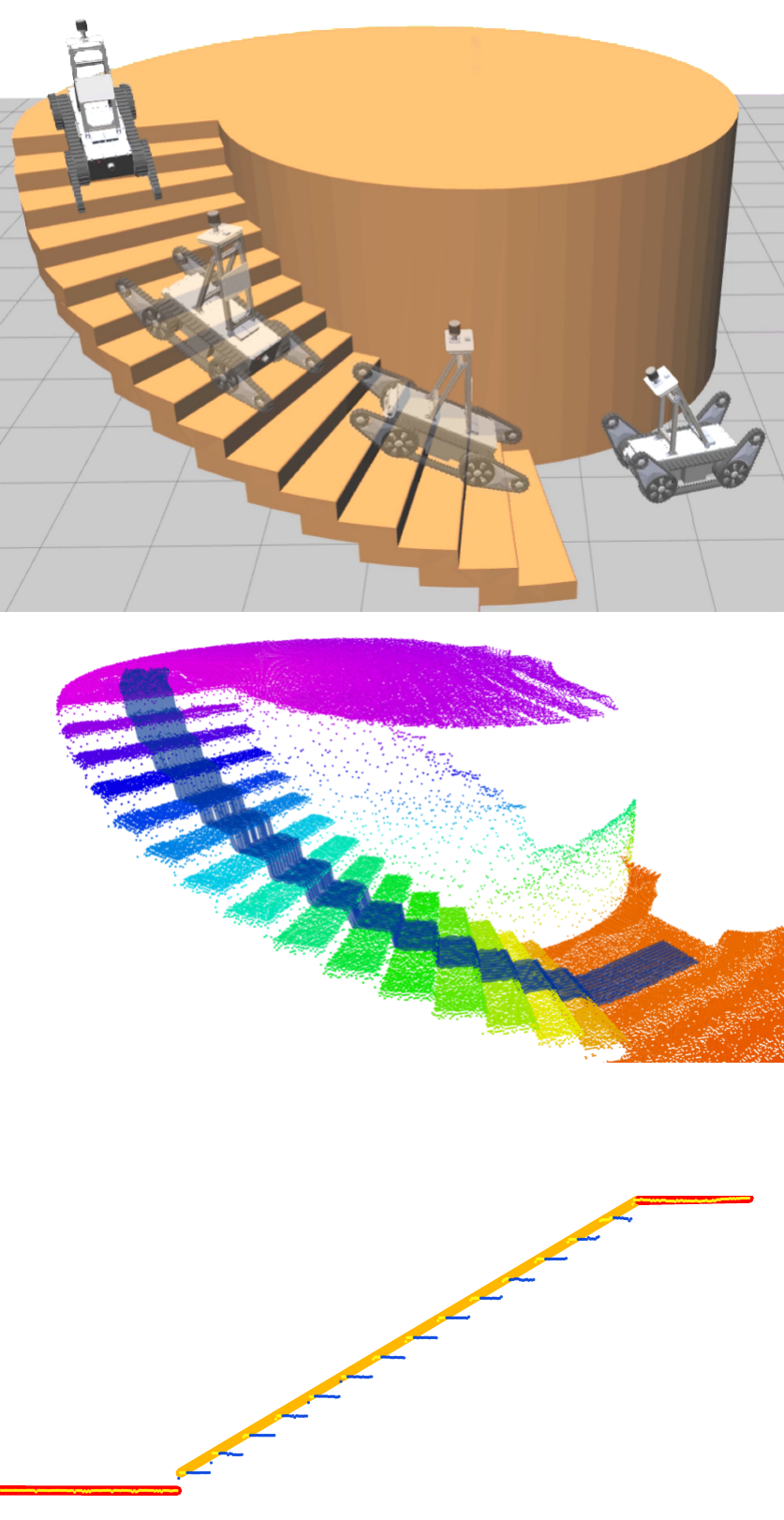}
    \label{fig:spiral_stairs}
    }
    \vfill
    \vspace{-0.4cm}
    \subfloat{
    \centering
        \hspace{-0.2cm}
        \includegraphics[width = 1.0\linewidth]{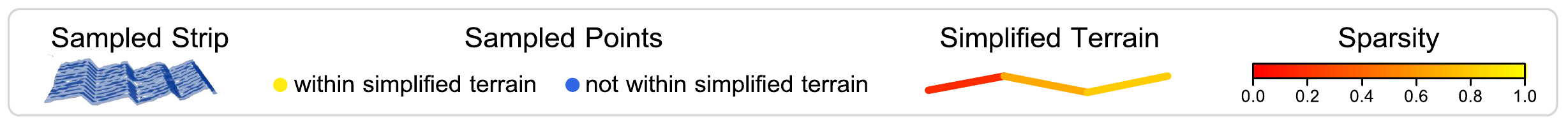}
        \label{fig:legend_for_simulation}
    }
    \vspace{-0.1cm}
    \caption{
    Snapshot of the simulation experiments and results of map sampling and terrain simplification.  (a) A high platform of 0.4m in height and 1.2m in length; (b) Straight stairs ascending and descending with each step 0.2m in height and 0.3m in width; (c)  Spiral stairs with a radius of 3.8m, each step 0.2m in height and 0.38m in average width. The high platform and straight stairs are used to evaluate the effectiveness of the method in traversing obstacles with heights close to the flipper length and on sloping, sparse terrain, respectively. Spiral stairs are used to demonstrate that our RTC model and methods are still effective on a curved reference path.}
    \label{fig:simulation_snapshot}
    \vspace{-0.4cm}
\end{figure}

The experiment aims to demonstrate the advantages of our approach in terms of time and energy efficiency, stability, and smoothness.
We conduct terrain traversal comparative experiments in three typical scenarios shown in Fig. \ref{fig:simulation_snapshot}. 
Our approach is compared with expert operator control, which serves as a widely accepted baseline for evaluation \cite{zimmermann2014adaptive,gianni2016adaptive,kojima2020stable}, as well as a best-effort reimplementation of the state-of-the-art GFMP method \cite{chengeometry}.

Given that the GFMP method faces a trade-off between computational efficiency and motion quality, we employ two versions of it: 
(1) offline full horizon, high-resolution planning 
(covers entire reference path; waypoint resolution: 0.05m; flipper angle resolution: 1$\degree$) and 
(2) online limited horizon, low-resolution planning 
(waypoint number: 4; waypoint resolution: 0.2m; flipper angle resolution: 10$\degree$).
Parameters for both versions are based on the specifications outlined in \cite{chengeometry} and tuned to perform best.
Notably, the offline version, although not explicitly proposed in the paper, significantly outperforms the online version.

To quantitatively compare the traversal quality, five metrics are proposed and evaluated:

(\romannumeral1) T: total traversal time, an indicator of time efficiency;

(\romannumeral2) RA$=\int_{0}^{T}(|\dot{\theta}_{f}|+|\dot{\theta}_{r}|) \mathrm{d} t$: accumulated absolute rotation angle of flippers. With the same driving distance in the same scenario, the difference in energy consumption is reflected in the flipper motion, so RA is a proxy for energy efficiency;

(\romannumeral3) MP: maximum absolute pitch angle. The larger its value, the more likely the robot will tip over;

(\romannumeral4) MPA: maximum absolute pitch angle acceleration. It reflects the excessive shock suffered by the robot;

(\romannumeral5) RMSA$=\!\mathrm{rms}( \ddot{\theta}_f ) \!+\!\mathrm{rms}( \ddot{\theta}_r ) $: root mean squared of flipper angular acceleration, reflecting the motion smoothness.

We conduct five trials with each method in each scenario, setting the identical initial and goal positions and the same maximum track and flipper velocity settings.
The results of the average metrics are presented in Table \ref{tab:result_table}.
\begin{table}[!htbp]
    \setlength\tabcolsep{4pt}
    \centering
    \renewcommand\arraystretch{1.3}
    \caption{Results of Typical Terrain Traversal Quality of Different Methods}
    \resizebox{1.0\linewidth}{!}{ 
    \begin{tabular}{|c|c|c|c|c|c|c||c|}
    \hline
    Scene   & Method & \makecell{T\\($\mathrm{s}$)} & \makecell{RA\\($\degree$)} & \makecell{MP\\($\degree$)} & \makecell{MPA \\ ($\degree/\mathrm{s}^2$)} & \makecell{RMSA\\ ($\degree/\mathrm{s}^2$)} & Score \\
    \hline
    
    \multirow{4}{*}{\makecell{High\\Platform}} 
    & Operator & \textbf{43.73} & 560.38  & 19.29  & \underline{83.90} & \underline{33.95} & \underline{0.2935} \\
    & GFMP (offline) & \underline{44.42} & \underline{522.38}    & \textbf{11.44}   & 138.92 & 48.34 & 0.4117 \\
    & GFMP (online) & 45.15 & 745.28  & 17.15   & 147.86 & 121.83 &  0.9291 \\
    & HTO (ours) & 45.61  & \textbf{484.26}  & \underline{16.86}    & \textbf{48.20} & \textbf{28.50} & \textbf{0.1650} \\ \hline
    
    \multirow{4}{*}{\makecell{Straight\\Stairs}} 
    & Operator & 75.85 & 721.15  & \underline{35.47}  & 93.50 & \underline{31.27} & 0.1809 \\
    & GFMP (offline) & \underline{75.13} & \underline{675.42} & 37.33  & \textbf{43.58} & 44.72 & \underline{0.1336} \\
    & GFMP (online) & 75.61  & 1110.45  & 37.73  & 259.62 & 104.81 & 0.9984 \\
    & HTO (ours)  & \textbf{69.32}  & \textbf{617.06} & \textbf{35.18}   & \underline{56.31} & \textbf{28.65} & \textbf{0.0428} \\ \hline
    
    \multirow{4}{*}{\makecell{Spiral\\Stairs}}
    & Operator & 104.46 & \underline{566.08}  & \underline{32.34}  & 121.39 & \textbf{22.17} & 0.1537 \\
    & GFMP (offline) & 111.83 & 567.64  & 32.74  & \underline{92.32} & 33.47 & \underline{0.1379} \\
    & GFMP (online) & \underline{94.88}  & 1613.08  & 35.54  & 142.36 &  136.21 & 0.9230 \\
    & HTO (ours)  & \textbf{87.51}  & \textbf{522.18}   & \textbf{31.90}  & \textbf{86.67} & \underline{24.10} & \textbf{0.0131} \\ \hline
    \multicolumn{8}{p{\dimexpr\linewidth+20\tabcolsep}}{\textit{Note}: \textbf{Bold} indicates the best performance and \underline{underline} indicates the second best. Evaluation scores range from 0 to 1, calculated using the TOPSIS method, where a smaller score indicates better performance.}
    \end{tabular}
    }
    \label{tab:result_table}
    \vspace{-0.2cm}
\end{table}

The expert operator is pretrained to reduce unnecessary flipper adjustments during traversal. However, their time efficiency is relatively low on stairs due to their prioritized adjustment of the flippers to ensure stability. In addition, the analog input of the joystick improves motion smoothness.

For the GFMP method, the offline version performs well in most metrics, which is attributed to the high resolution and full horizon planning. 
However, this comes at the cost of longer computation times, which take tens of minutes per computation, compared to only 150 ms for ours. 
It is important to emphasize that these extended computation times are simply unacceptable for physical robots in complex environments. 
GFMP aims to minimize flipper adjustments, resulting in a small RA in the offline version, similar to our approach. 
However, the online version is restricted by a limited planning horizon, neglecting overall motion efficiency and resulting in the largest RA due to unnecessary flipper motion. 
In addition, the online version often fails to ensure the stability of transitions between planning waypoints due to low waypoint resolution, leading to tipping over. Such incidents can cause excessive shocks on the joints, leading to damage to the mechanism and motors. 

In contrast, our HTO method excels in overall motion quality. 
It demonstrates significantly shorter traversal times on stairs and possesses the highest energy efficiency. 
Also, it performs notably well in terms of stability with minimal occurrences of excessive shocks. 
Moreover, compared to GFMP, the Hermite interpolation we employ generates a smoother flipper motion.

\subsection{Evaluation of Multi-Objective Cost Function}
\label{subsec:simulation_cost_experiment}
To evaluate the effectiveness of each cost term, we conduct comparative experiments on straight stairs by removing each term. 
After removing the time cost, the robot can not accelerate and remain stationary, as minimizing the time cost incentive drives the robot forward.
The five-trial average of the aforementioned five metrics is obtained by removing the other two cost terms, as shown in table \ref{tab:cost_result_table}.
\begin{table}[!htbp]
    \setlength\tabcolsep{4pt}
    \centering
    \renewcommand\arraystretch{1.3}
    \caption{Effects of Cost Terms on the Straight Stairs}
    \vspace{-0.2cm}
    \resizebox{0.8\linewidth}{!}{ 
    \begin{tabular}{|c|c|c|c|c|c|}
    \hline
    Cost Term  & \makecell{T\\($\mathrm{s}$)} & \makecell{RA\\($\degree$)} & \makecell{MP\\($\degree$)} & \makecell{MPA \\ ($\degree/\mathrm{s}^2$)} & \makecell{RMSA\\ ($\degree/\mathrm{s}^2$)} \\ \hline
    w/o $J_{coh}$ & 72.62 & 640.79 & 36.71 & 88.48 &  29.04\\ 
    w/o $J_{stab}$   & 84.23 & 659.47 & 35.00  & 76.13 & 30.98 \\ 
    w/o $J_{coh}$, $J_{stab}$ & 80.34 & 919.12 & 36.99 & 125.68 & 35.55 \\ \hline
    \end{tabular}
    }
    \vspace{-0.2cm}
    \label{tab:cost_result_table}
\end{table}

When the coherence cost is removed, the robot occasionally exhibits unnecessary back-and-forth flipper motion, thereby increasing both the T and RA. 
When the stability cost is removed, the robot is no longer flat on the simplified terrain of the stairs.
Instead, the extra motion of the flippers brings their endpoints to contact with the steps.
This not only increases the T and RA compared to the full cost function in Table \ref{tab:result_table}, but it can also result in the robot getting stuck during ascent and falling vertically during descent.
When both stability and coherence costs are removed, insufficient support leads to poorer stability, 
the extra flipper motion leads to a longer traversal time, and the overall performance metrics drop significantly.
Its traversal quality significantly diminishes, but overall it is still superior to GFMP (online). 
The results of RMSA show that smoothness is still guaranteed by the interpolated trajectories. 
In conclusion, the experiments demonstrate that time cost is an indispensable cost term, while stability and coherence cost can significantly improve the time, energy efficiency, and stability of traversal. 

\begin{figure*}[!t]
    \centering
    \includegraphics[width = 1.0\linewidth]{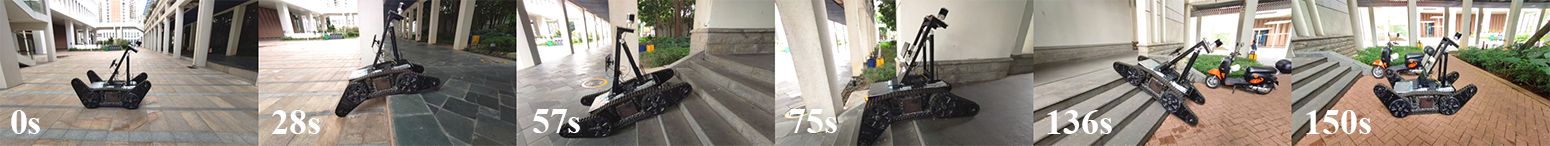}
    \vspace{-0.55cm}
    \caption{
    Snapshot of one real-world experiment with multiple terrains, with a total distance of about 24m. 
    The robot needs to first pass through a platform with a height of 0.3m (28s), then traverse six steps (0.16m high and 0.3m wide) (57s and 75s), and climb down four steps of the same size (136s) to reach the ground (150s) after turning in the corridor.
    }
    \vspace{-0.3cm}
    \label{fig:real_world_snapshot}
\end{figure*}

\begin{figure*}[!t]
    \centering
    \includegraphics[width = 1.0\linewidth]{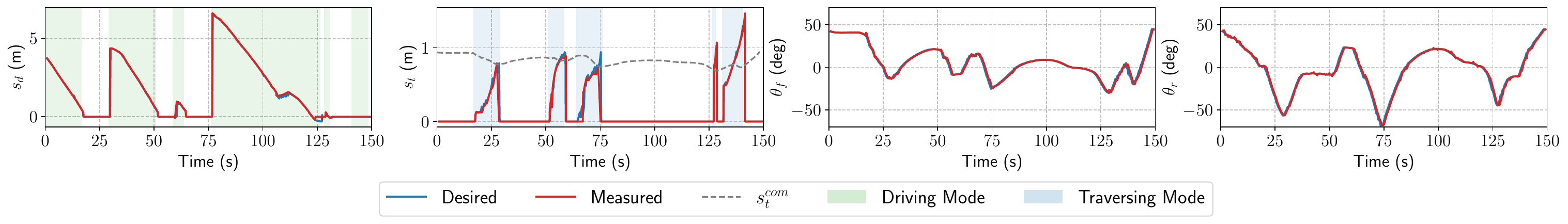}
    \vspace{-0.55cm}
    \caption{
    Planned trajectory $\left\{ s_d(t),s_t(t),\theta _f(t),\theta _r(t) \right\}$ and the corresponding measurement for the traversing mode period and driving mode period are shaded in blue and green, respectively (the measured $s_t$ for driving and $s_d$ for traversing are set to 0). 
    When $s_t$ surpasses $s_t^{com}$, the robot achieves stability on the subsequent terrain. 
    During the 108-113s when the robot is tracking the curved path, there is a brief increase in $s_d$ due to the tracking error.
    The last driving mode period has no $s_d$ since the robot has already reached the goal terrain and decelerated without planning.
    }
    \vspace{-0.4cm}
    \label{fig:real_world_plot}
\end{figure*}

\vspace{-0.1cm}
\subsection{Multi-Terrain Traversal in Real-World Scenarios}
\label{subsec:Real Scenario Experiments}
We deploy the terrain traversal system on the Searcher robotic platform to conduct experiments in real-world scenarios, as shown in the supplemental video, to demonstrate that our approach can traverse multiple terrains in succession with receding-horizon planning.
Fig. \ref{fig:real_world_snapshot} shows a collection of snapshots from one of the experiments, a scenario involving a high platform and staircases.
As expected, the HTO module successfully generates the hybrid motion and the controller can track the reference trajectory, as shown in Fig.\ref{fig:real_world_plot}.

\subsubsection{Results of Mode Switching}
Fig.\ref{fig:real_world_plot} shows that the HTO module is capable of generating motion with mode switching in a multi-terrain traversal.
The whole traversal task contains 10 mode switches, which are efficiently solved by dividing into subproblems with single mode switches in a receding-horizon fashion.
Specifically, the robot successfully climbs up the high platform by switching to traversing mode within 18-29s.
Each staircase's traversal undergoes two traversal modes and one driving mode since the steps are simplified to a plane and the planner regards the robot as driving on a plane when the body is entirely situated on it.
A short driving mode occurs during 129-131s due to the stairs' brief length.

\subsubsection{Tracking Performance}
The planned trajectory and measurements show that the controller has excellent tracking performance, as evidenced by the mean absolute tracking errors with standard deviations along the entire traversal, which are as follows: $s_d$: $3.09\pm5.80$ (cm), $s_t$: $2.74\pm4.28$ (cm), $\theta_f$: $1.31\pm1.96$ ($\degree$), and $\theta_r$: $1.78\pm2.40$ ($\degree$).
\section{Conclusion and Future Work}
\label{sec:conclusion}
To the best of our knowledge, this work is the first application of TO techniques to articulated tracked robots for terrain traversal.
We propose a planar RTC model to describe configuration and facilitate dimensionality reduction.
A novel HTO formulation with a multi-objective cost function is presented, divided into subproblems 
to be solved in real-time with a receding horizon to generate motions with hybrid mode switching.
In addition, a terrain traversal system containing map sampling, terrain simplification, HTO, and a 
tracking controller is designed and deployed.
The effectiveness of our approach is verified on simulated and physical platforms and demonstrated through comparative experiments showing advantages over expert operator control and the state-of-the-art method in terms of time and energy efficiency, stability, and motion smoothness.
Future work will focus on global path planning based on online mapping to further improve system autonomy to perform full-stack autonomous navigation and complex tasks.
\bibliographystyle{IEEEtran}
\renewcommand{\baselinestretch}{0.842}
\bibliography{IEEEabrv,bib/bibliography}
\end{document}